\relax
\documentclass[letterpaper]{article} 
\usepackage{aaai19}  
\usepackage{times}  
\usepackage{helvet}  
\usepackage{courier}  
\usepackage{url}  
\usepackage{graphicx}  
\usepackage{amsmath}
\usepackage{subcaption}
\usepackage{enumerate}
\usepackage{booktabs}
\usepackage{multirow}
\usepackage{bm}
\usepackage{amssymb}
\begin{document}

\title{Hybrid Self-Attention Network for Machine Translation}
\author{$^1$Kaitao Song, $^2$Xu Tan, $^3$Furong Peng, $^1$Jianfeng Lu\\
$^1$Nanjing University of Science and Technology \\
$^2$Microsoft Research \\
$^3$Institute of Big Data Science and Industry, Shanxi University \\
{\tt \small \{kt.song, lujf\}@njust.edu.cn} \qquad
{\tt \small xuta@microsoft.com}  \qquad
{\tt \small pengfr@sxu.edu.cn}
}

\maketitle
\begin{abstract}
The encoder-decoder is the typical framework for Neural Machine Translation (NMT), and different structures have been developed for improving the translation performance. Transformer is one of the most promising structures, which can leverage the self-attention mechanism to capture the semantic dependency from global view. However, it cannot distinguish the relative position of different tokens very well, such as the tokens located at the left or right of the current token, and cannot focus on the local information around the current token either. To alleviate these problems, we propose a novel attention mechanism named Hybrid Self-Attention Network (HySAN) which accommodates some specific-designed masks for self-attention network to extract various semantic, such as the global/local information, the left/right part context. Finally, a squeeze gate is introduced to combine different kinds of SANs for fusion. Experimental results on three machine translation tasks show that our proposed framework outperforms the Transformer baseline significantly and achieves superior results over state-of-the-art NMT systems.
\end{abstract}

\section{Introduction}
Recently, Neural Machine Translation (NMT) has witnessed a rapid and revolutionary change in the development of sequence transduction model. Recurrent Neural Networks (RNNs), especially like long short term memory \cite{Hochreiter:1997:LSM:265493.264179}, has achieved some promising performance in machine translation \cite{kalchbrenner13emnlp,43155}. However, recurrent neural network depends on the previously hidden state that cannot support parallel computation efficiently. In order to address this problem, researchers have concentrated on leveraging parallelizable model structure into NMT. Convolutional neural networks (CNNs) are first introduced into NMT and achieved promising performance 	\cite{DBLP:journals/corr/KalchbrennerESO16,gehring2017convs2s}.

To improve the alignment accuracy from target sentence to source sentence, attention mechanism has been applied to NMT \cite{bahdanau+al-2014-nmt} and becomes an indispensable part in NMT framework. Attention model can help the decoder to capture a global context abstraction of entire source sentence and learn which source word should be aligned. Inspired by this property of attention mechanism, Vaswani \shortcite{46201} proposed a novel sequence-to-sequence framework called \emph{Transformer}. Transformer is an encoder-decoder structure that is only composed of a novel network named as self-attention. Self-attention network (SAN) can extract context-aware features inside the encoder or decoder based on a scaled multi-head attention mechanism. SAN supports highly parallelizable computation. And more important, it enables better representation of learning long range dependency due to the shorter distance between any two elements.

\begin{figure}
	\centering	
	\includegraphics[width=0.75\columnwidth, height=25ex]{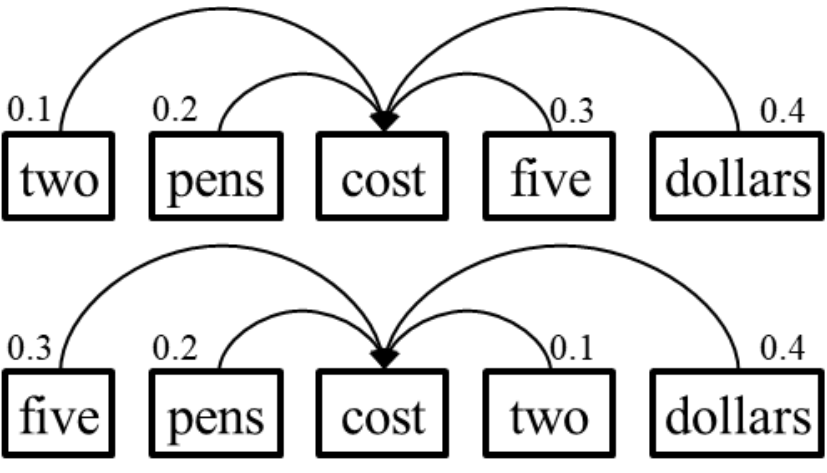}
	\caption{\label{introduction} Two sentences that ``five" and ``two" exchange order. The number over the tokens is relationship ratio.}
\end{figure}

However, there still exists some insufficiencies in this attention mechanism. Attention mechanism calculates the relationship between each position by a weighted average. In other words, the calculation of SAN lacks temporal order information which is necessary for modeling sequence. For example in Figure \ref{introduction}, token ``\emph{cost}'' will collect the same information from sentence ``\emph{two pens cost five dollars}'' and ``\emph{five pens cost two dollars}''. But the content of these two sentences should be distinct. Compared with SAN, convolution network provides a linear transformation from different positions and recurrent network directly relies on the previous state. That is why Transformer needs to provide positional embedding for SAN. Therefore, SAN is sensitive to positional embedding and how to model relative position information for SAN is still a troublesome problem. 

In addition, we think that extracting partial information can also provide the model a multidimensional understanding from the local context, such as convolution. We also take Figure \ref{introduction} as an example: ``\emph{two}" and ``\emph{five}" can be regarded as the modifiers of ``\emph{pens}" and ``\emph{dollars}" respectively that affects the meaning of the sentence, while the conventional SAN does not possess such relationship. In other words, the model is able to obtain the benefit if we can extract semantics from local context and provide it to SAN.

In brief, the above issues are the core problems to be targeted in this paper. As is known, allowing the network to extract multi-scale features is a superior technique in the field of computer vision such as GoogleNet \cite{43022}. So we have a thinking, is it possible to let SAN abstract deep representations from different levels rather than only the global context?

In this paper, we propose a novel self-attention architecture to produce plentiful representations from different aspects for machine translation. We name our attention model as ``Hybrid Self-Attention Network (HySAN)" which can extract temporal order information and partial information besides global information. Our proposed HySAN is also easy to implement, which just needs to add multiple branches after the dot product operation of the SAN module. Besides, our proposed method barely increases any other parameters. 

We evaluate HySAN on IWSLT14 German-English, WMT14 English-German and WMT17 Chinese-English translation tasks, respectively. Experimental results indicate that our method outperforms baseline by 1.0 BLEU, 0.6 BLEU, 0.4 BLEU and 1.07 BLEU score in IWSLT14 German-English (small setting), WMT14 English-German (base setting), WMT14 English-German (big setting) and WMT17 Chinese-English (big setting) respectively.


\section{Background}

\subsection{Encoder-Decoder based NMT}
Given a source sentence $\mathbf{x}=\{x_1,\ldots,x_n\}$, and a generated target sentence   $\mathbf{y}=\{y_1,\ldots,y_m\}$, where ${x_i}$ and ${y_t}$ are the $i$-th and $t$-th token for source and target sentence respectively, $n$ and $m$ are the length of sentence $\mathbf{x}$ and $\mathbf{y}$. The NMT model learns to maximize the log likeihood function $\mathrm{P}(\mathbf{y}|\mathbf{x})$. And $\mathrm{P}(\mathbf{y}|\mathbf{x})$ is calculated as:
\begin{equation}
\mathrm{P}(\mathbf{y}|\mathbf{x})=\prod_{t=1}^m \mathrm{P}(y_t|y_{<t},x;\theta),
\end{equation}
where $\theta$ is the model parameter. The probability of generating next target word $y_i$ is:
\begin{equation}
\mathrm{P}(y_t|y_t,\mathbf{x};\theta) \propto \exp\{f(y_{<t},c_i;\theta)\},
\end{equation}
where $c_i$ is the compressed source representation for generating the $i$-th target word, and $f(\cdot)$ is model function of the decoder. In our work, $f(\cdot)$ refers to self-attention network.

Attention mechanism is usually used to measure the matching degree between source word and target word.  Generally, there exists several alternatives in calculating alignment for attention mechanism, such as \emph{dot}, \emph{concat} and \emph{general} \cite{DBLP:conf/emnlp/LuongPM15,bahdanau+al-2014-nmt}. Following Vaswani \shortcite{46201}, our work adopt \emph{dot} as the default form which can be described as a query-key-value form:
\begin{equation}
\label{attention}
\begin{aligned}
\mathrm{Attention(Q, K, V)} &= \mathrm{softmax(QK^T)V}.
\end{aligned}
\end{equation}
More specifically, it can be regarded as:
\begin{equation}
\begin{aligned}
\mathrm{Attention(Q, K, V)} &= \mathrm{\left[\sum_{i=1}^{n} \alpha(q_t, k_i) \cdot v_i \right]_{t=1}^m}, \\
\mathrm{\alpha(q_t, k_i)} &= \mathrm{\frac{\exp(q_t \cdot k_i)}{\sum_{j=1}^{n}\exp(q_t \cdot k_j)}},
\end{aligned}
\end{equation}
where $\mathrm{Attention}(\cdot,\cdot,\cdot)$ is mapped to the function of an attention model, $\mathrm{q_t}$ usually represents the hidden state of the $\mathrm{t}$-th token, and $\mathrm{\alpha(\cdot,\cdot)}$ is used to calculate the matching degree between the $\mathrm{j}$-th token in source sentence and $\mathrm{t}$-th token in target sentence. $\mathrm{v_i}$ and $\mathrm{k_i}$ refer to the $\mathrm{i}$-th hidden representation of the source sentence. 

\begin{figure*}
	\centering
	\includegraphics[width=2\columnwidth, height=60ex]{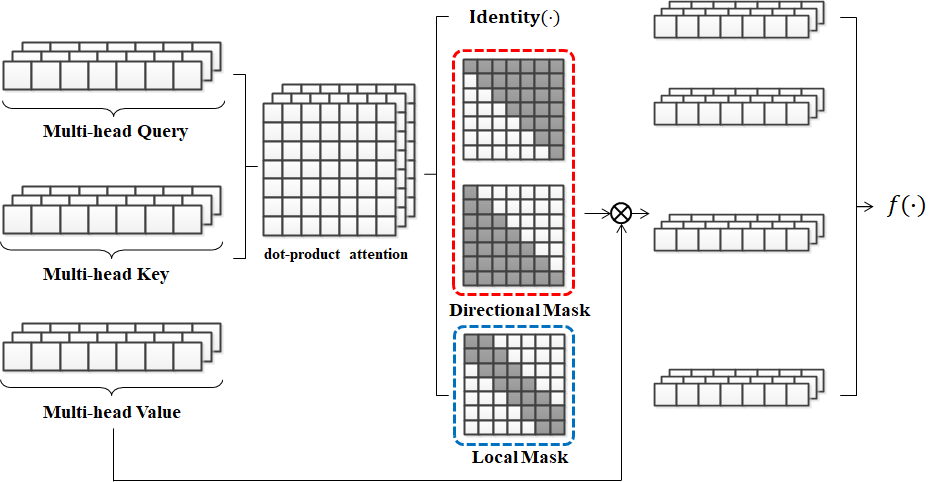}
	\caption{\label{arch} The architecture of multi-head hybrid self-attention network. We deliver features from different branches into $f(\cdot)$ for combination. Red dotted box represents directional masked matrix, and blue dotted box is the local masked matrix.}
\end{figure*}

\subsection{Transformer Network}
Transformer mainly consists of a stack of attention layers, which is composed of a multi-head attention sub-layer and a feed-forward sub-layer. To enable deep network and normalization for neurons, Transformer also adds layer normalization \cite{DBLP:journals/corr/BaKH16} after each sub-layer and uses residual connections \cite{DBLP:conf/cvpr/HeZRS16} around each layer. 

In order to incorporate the temporal order information into the model, Transformer incorporates position embedding $\mathbf{p}=(p_1,\ldots,p_n)$ into learned word embedding $\mathbf{e}=(e_1,\ldots,e_n)$. The parameter of positional embedding is initialized as:
\begin{equation}
\begin{aligned}
\mathbf{P}(\mathrm{pos}, 2i)&=\sin(\mathrm{pos}/10000^{2i/\mathrm{d_{model}}}), \\
\mathbf{P}(\mathrm{pos}, 2i+1)&=\cos(\mathrm{pos}/10000^{2i/\mathrm{d_{model}}}),
\end{aligned}
\end{equation}
where $\mathrm{pos}$ is the position of the word, $i$ is the index of dimension and $\mathrm{d_{model}}$ is the dimensions of the embedding layer.

Multi-head attention mechanism is built upon scaled dot product attention. Multi-head attention mechanism obtains $h$ different representations of the (query, key, value) form, computes each representation and then concatenates. It can be formulated as follows:
\begin{equation}
\begin{aligned}
\mathrm{SA(q, k, v)} &= \mathrm{MH(q, k, v) W^o},\\
\mathrm{MH(q, k, v)} &= \mathrm{[H_1(q,k,v),...,H_h(q,k,v)]}, \\
\mathrm{H_i(q, k, v)} &= \mathrm{Attention(\frac{qW_i^q}{\sqrt{d_s}},kW_i^k,vW_i^v)},
\end{aligned}
\end{equation}
where $\mathrm{MH}(\cdot)$ represents the function of multi-head attention model, $\mathrm{H}_i(\cdot)$ means the output of single head where $h$ is the number of head. $\mathrm{Attention}(\cdot)$ is defined in equation \ref{attention}. $\mathrm{d_s}$ is the dimension of the query. $\mathrm{W^o}  \in \mathbb{R}^{\mathrm{h d_{v} \times d_{model}}}$, $\mathrm{W_i^q} \in \mathbb{R}^{\mathrm{d_{model} \times d_q}}$, $\mathrm{W_i^k} \in \mathbb{R}^{\mathrm{d_{model} \times d_k}}$, $\mathrm{W_i^v} \in \mathbb{R}^{\mathrm{d_{model} \times d_v}}$ are all parameter matrices. Noting that $\mathrm{d_q = d_k = d_{model}}$ in general. This architecture helps model to learn separate relationships from different heads. Besides, it is necessary to apply a mask for self-attention in the decoder structure to prevent position from attending to subsequent positions.

The following component after the multi-head attention is a feed-forward network, which is applied to each position separately and identically. This network is composed of two linear transformations with a ReLU activation \cite{DBLP:conf/icml/2010} in between:
\begin{equation}
\mathrm{FFN(x) = \max(0, xW_1 + b_1)W_2 + b_2},
\end{equation}
where $\mathrm{W_1}$ and $\mathrm{W_2}$ are weight parameters, $\mathrm{b_1}$ and $\mathrm{b_2}$ are bias parameters.

\section{Approach}
Self-attention network is an advanced model structure that can learn dependency from global context directly. In our paper, we apply two specific-designed attention masks into the self-attention network as the additional branches that support network to extract temporal order information and local information respectively. We call these branch networks as ``Directional Self-Attention Network (DiSAN)" and ``Local Self-Attention Network (LSAN)". We mix these different SAN branches to form a ``Hybrid Self-Attention Network (HySAN)" with a squeeze gate mechanism network. The detail of our model structure is illustrated in Figure \ref{arch}.


\subsection{Directional Self-Attention}
Directional self-attention (DiSAN) is a masked self-attention block to explore the directional and temporal order information. Self-attention learns the dependency of a token on all the tokens in the sentence purely based dot product attention, and treats each token equivalently. Consequently, the alignment score is the key point of attention mechanism because it contains all relationship between any two positions. Besides, the alignment matrix in SAN is a square matrix whose order is $\mathrm{n}$. Therefore, we design a masked matrix to achieve the goal:
\begin{equation}
\begin{aligned}
\mathrm{DiSAN(q, k)} &= \mathrm{softmax(qk^T + M) v}, \\
\mathrm{where} \; \mathrm{M} &= \mathrm{M_{fw} \; or \; M_{bw}} ,	
\end{aligned}
\end{equation}
while $\mathrm{M_{fw}}$ and $\mathrm{M_{bw}}$ are denoted as:
\begin{equation}
\begin{aligned}
\mathrm{M_{fw}(i, j)} &= 
\begin{cases}
0, & \mathrm{j <= i} \\
-\infty, & \mathrm{otherwise}
\end{cases}, \\
\mathrm{M_{bw}(i, j)} &=
\begin{cases}
0, & \mathrm{i <= j} \\
-\infty, & \mathrm{otherwise}
\end{cases} 
\end{aligned}
\end{equation}
Here $\mathrm{M_{fw}}$ means the forward masked matrix and $\mathrm{M_{bw}}$ means the backward masked matrix. In the forward direction of SAN, the token $\mathrm{i}$ can attend to the position $\mathrm{j}$ which is earlier than $\mathrm{i}$, and vice versa in the backward direction. DiSAN helps the model to abstract semantics with temporal order information and context-aware representation. Note that DiSAN can only be applied for encoder structure, because the decoder of the S2S structure has been restricted to view only forward direction.

\subsection{Local Self-Attention}
Self-attention usually focuses on the global dependency while ignores some local information. As introduced in the above section, the alignment matrix can be viewed as a $\mathrm{n \times n}$ relationship feature matrix. Compared to SAN, convolution network usually creates representations in a local context with the fixed size. Inspired by this property, we design a symmetrical masked matrix for SAN to collect features in local context. Similar to the local connection in convolution network, LSAN can view $\mathrm{2k+1}$ elements in a sliding window if the radial width of the window is $\mathrm{k}$. The LSAN is formulated as:
\begin{equation}
\begin{aligned}
\mathrm{LSAN(q,k,v)} &= \mathrm{softmax(qk^T + M) v} \\
\mathrm{where \; M} &= \mathrm{M_{e} \; or \; M_{d}}		
\end{aligned},
\end{equation}
where $\mathrm{M_e}$ and $\mathrm{M_d}$ represent the mask applied to the encoder and decoder respectively. $\mathrm{M_e}$ and $\mathrm{M_d}$ are denoted as:
\begin{equation}
\begin{aligned}
\mathrm{M_{e}(i, j)} &= 
\begin{cases}
0, & \mathrm{|i-j| \leq k} \\
-\infty, & \mathrm{otherwise}
\end{cases}, \\
\mathrm{M_{d}(i, j)} &=
\begin{cases}
0, & \mathrm{i - j \leq k} \\
-\infty, & \mathrm{otherwise}
\end{cases}.
\end{aligned}
\end{equation}
The self-attention equipped with this designed mask matrix can abstract information from local context. However, it is inefficient to only deploy local self-attention because it contains large unused elements-wise operations. Therefore, we consider the local self-attention as an additional branch to share the results of the dot product with global self-attention. By this sharing operation, our method will not produce any redundant computation and any additional parameters except for the fusion stage. Besides, our method will not cause patch alignment issues that usually occurred in the convolution network. In our work, we only consider $\mathrm{k=1,2,5}$ for ablation studies.

\begin{figure}
	\centering
	\includegraphics[width=0.5\columnwidth, height=35ex]{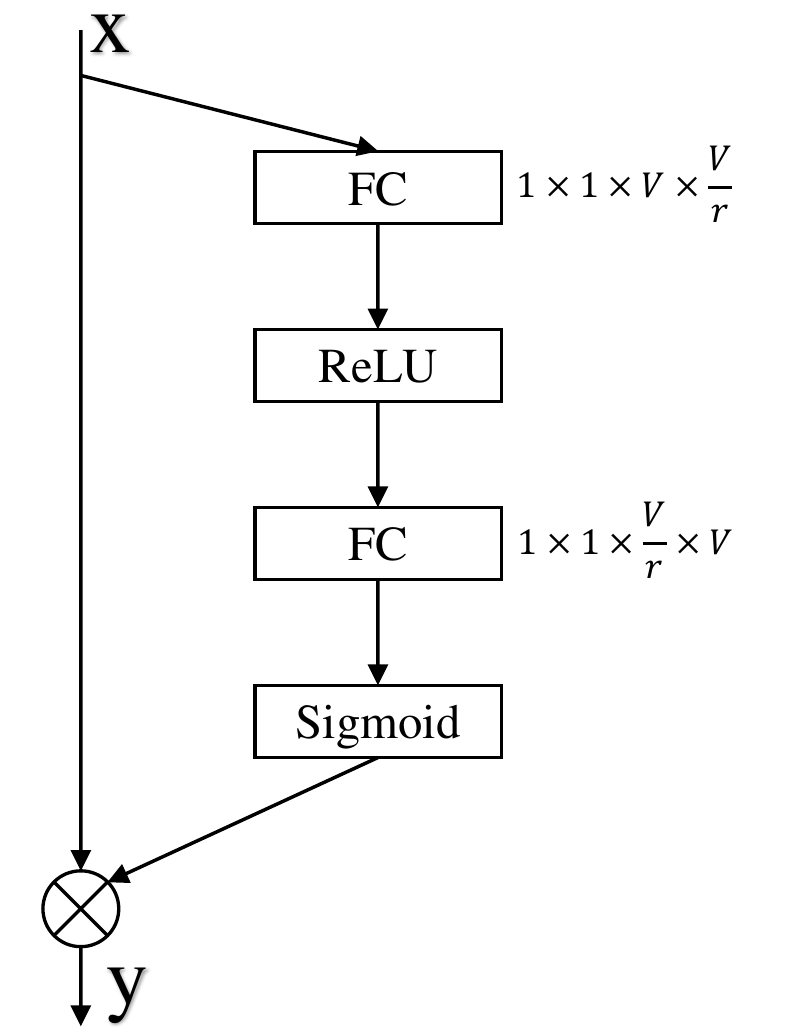}
	\caption{\label{gate}The schema of the gate module for single branch. $V$ means the dimension of the \emph{Value} in attention model. $r$ is a scaled factor. $\otimes$ equals multiplication.}
\end{figure}

\subsection{Fusion}
Following the operation of each SAN branch, we rewrite the output of HySAN as:
\begin{equation}
\begin{aligned}
\mathrm{out(q, k, v)} &=\mathrm{f(\left[softmax(qk^T + M_i)v\right]_{i=1}^\mathit{l}}), \\
\end{aligned}
\end{equation}
here $\mathrm{M_i}$ refers to the mask matrix of DiSAN, LSAN or Global SAN, $\mathit{l}$ means the number of branches and $\mathrm{f(\cdot)}$ represents fusion function. Generally, there are many fusion methods for candidates. In order to make use of the aggregated information, we consider three different alternatives for aggregation here:
\begin{equation}
\label{typeofconcat}
\mathrm{f(x)} = 
\begin{cases} 
\sum_1^\mathit{l}x_i,  & \mathrm{Sum} \\
\mathrm{H}([x_1,...,x_{\mathit{l}}]), & \mathrm{Concat} \\
\sum_1^\mathit{l}x_i * \mathrm{SG}(x_i), & \mathrm{Gated \; Sum}
\end{cases}
\end{equation}
where $\mathrm{H}(\cdot)$ is a fully connected function for dimension scaling, and $\mathrm{SG}(\cdot)$ is a squeeze gate mechanism that is shown in Figure \ref{gate}. The squeeze gate network is comprised of two modules: a squeeze block and a gate block. The squeeze block consists of two fully connected networks with a ReLU activation function in between. The purpose of the first linear layer is to reduce dimension while the second one is to increase dimension. The gate block is designed as a gate mechanism with a sigmoid activation function. The output of the gate block represents the scale of this branch. $\mathrm{SG}(\cdot)$ can be simplified by:
\begin{equation}
\mathrm{SG(x) = \sigma(f_2(ReLU(f_1(x)))},
\end{equation}
here $\sigma(\cdot)$ is sigmoid activation function, $\mathrm{f_1(\cdot)}$ and $\mathrm{f_2(\cdot)}$ are feed-forward network. Figure \ref{gate} is a detailed illustration of the squeeze gate mechanism. The squeeze gate mechanism has two advantages: the first is to reduce parameters, and the second is to increase non-linearity. In the following section, we will explore the profit of different fusion methods for the model.

\begin{table*}[t]
	\centering
	\begin{tabular}{c|c c c c c c | c c c c | c | c | c | c }
		\hline
		\multirow{2}{*}{ID} & \multicolumn{6}{c|}{Encoder} & \multicolumn{4}{c|}{Decoder} & \multirow{2}{*}{BLEU} & \multirow{2}{*}{$\Delta$} & \multirow{2}{*}{NoPos} & \multirow{2}{*}{$\Delta$} \\
		& Global & FW & BW & $L_1$ & $L_2$ & $L_5$ & FW & $L_1$ & $L_2$ & $L_5$ & & & &\\
		\hline
		$base$ & $\surd$ &  &  &  & &  & $\surd$ & &  & &  $31.27$ & - & $15.56$ & - \\
		\hline
		$a_1$ & $\surd$ & $\surd$ &  &  & &  & $\surd$ & &  & &  $31.50$ & $0.23$ & $28.78$ & $13.22$\\
		$a_2$ & $\surd$ &  & $\surd$ &  & &  & $\surd$ & &  & &  $31.83$ & $0.56$ & $29.57$ & $14.01$\\
		$a_3$ & $\surd$ & $\surd$  & $\surd$ &  & &  & $\surd$ & &  & &  $31.87$ & $0.60$ & $30.01$ & $14.45$ \\
		\hline
		$b_1$ & $\surd$ &  &  & $\surd$ & &  & $\surd$ & &  & &  $31.55$ & $0.28$ & $27.65$ & $12.09$ \\		
		$b_2$ & $\surd$ &  &  &  & $\surd$ &  & $\surd$ & &  & &  $31.64$ & $0.37$ & $27.68$ & $12.12$ \\
		$b_3$ & $\surd$ &  &  &  &  & $\surd$ & $\surd$ & &  & &  $31.92$ & $0.65$ & $27.67$ & $12.11$ \\
		$b_4$ & $\surd$ &  &  & $\surd$ & $\surd$ & $\surd$ & $\surd$ & &  & & $31.85$ & $0.58$ & $29.87$ & $14.31$\\
		\hline
		$c_1$ & $\surd$ & $\surd$  &  & $\surd$  & &  & $\surd$ & &  & &  $31.77$ &$0.50$ & $30.40$ & $14.84$ \\
		$c_2$ & $\surd$ & $\surd$ &$\surd$ & $\surd$ & & & $\surd$ & &  &   &  $31.96$ & $0.69$ & $30.70$ & $15.14$ \\
		$c_3$ & $\surd$ &  & $\surd$ & & & $\surd$  & $\surd$ & &  & &  $32.02$  & $0.75$ & $30.34$ & $14.78$ \\
		$c_4$ & $\surd$ & $\surd$ &$\surd$ & & &  $\surd$& $\surd$ & &  &   &  $32.09$ & $0.82$ & $30.62$ & $15.06$ \\
		$c_5$ & $\surd$ & $\surd$ &$\surd$ & $\surd$ & $\surd$ & $\surd$ & $\surd$ & &  &   &  $32.16$ & $0.89$ & $30.68$& $15.12$ \\
		\hline
		$d_1$ & $\surd$ &  &  &  &  & & $\surd$ & $\surd$ &  & &  $31.34$ & $0.07$ & $15.61$ & $0.06$ \\
		$d_2$ & $\surd$ &  &  &  &  & & $\surd$ &  & $\surd$ & &  $31.30$ & $0.03$ & $15.62$ & $0.07$ \\
		$d_3$ & $\surd$ &  &  &  &  & & $\surd$ & &  &$\surd$   & $31.43$ & $0.16$ & $15.58$ & $0.03$ \\
		$d_4$ & $\surd$ &  &  &  &  & & $\surd$ & $\surd$ & $\surd$ & $\surd$ &  $31.49$ & $0.22$ & $15.70$& $0.15$ \\
		\hline
		$e_1$ & $\surd$ & $\surd$ & $\surd$ & & & $\surd$ & $\surd$ & & & $\surd$ &  $32.28$ & $1.01$ & $30.69$ & $15.13$ \\
		$e_2$ & $\surd$ & $\surd$ & $\surd$ & $\surd$ & $\surd$ & $\surd$ & $\surd$ & $\surd$& $\surd$ & $\surd$ &  $32.29$ &$1.02$ & $30.72$ & $15.16$\\
		\hline
	\end{tabular}

	\caption{\label{branch}Ablation studies of our method on IWSLT2014 German-English about the effect of branch. $L_k$ represents local SAN with width $k$. $\{a-e\}_{i}$ represents the ID in different experiment groups. ``$\Delta$" means increment on BLEU score. ``NoPos" represents BLEU score when disabling positional embedding.}
\end{table*}

\section{Experiment setting}
In this section, we will introduce the settings for experiments, including datasets, model settings and training details.
\subsection{Datasets}
Our approach is evaluated on a small translation task and two large translation tasks, which are IWSLT14 German-English, WMT14 English-German and WMT17 Chinese-English respectively.

\subsubsection*{IWSLT14 German-English}
For IWSLT14 German-English machine translation task \cite{Cettolo2014Report}, we tokenize the data  which comes from TED and TEDx talks, and remains 160K training sentences and 7K development sentences. We concatenate dev2010, tst2010, tst2011 and tst2012 as the test set.  We preprocess corpus with a 32000 word-piece \cite{DBLP:journals/corr/WuSCLNMKCGMKSJL16} shared vocabulary.

\subsubsection*{WMT14 English-German}
WMT14 English-German dataset \cite{Buck-commoncrawl} comprises about 4.5 million sentence pairs that are extracted from three corpora: Common Crawl corpus, News Commentary and Europarl v7. We adopt newtest2013 as the validate set and newtest2014 as the test set. The tokens are split with a 32000 word-piece shared vocabulary.

\subsubsection*{WMT17 Chinese-English}
WMT17 Chinese-English translation task contains 24M bilingual data, including News Commentary corpus, UN Parallel corpus and CWMT corpus. We use the same data selection as introduced in Hassan \shortcite{Hassan2018}. We keep 18M sentences after preprocessing. Newsdev2017 is used as the development set and newstest2017 is as the test set.

\subsection{Setup}
For IWSLT14 German-English translation task, the model is set as 2 hidden layers with 256 dimensions. And this task is deployed on single NVIDIA Titan X Pascal GPU for 100K steps. For WMT14 English-German translation task, we adopt 6 identical hidden layers with 512 dimensions for the base model, and 1024 dimensions for the big model. For WMT17 Chinese-English, we only use a big model setting. The number of the attention head of small, base and big model settings are set as 4, 8 and 16 respectively. We use 8 NVIDIA P100 GPUs for WMT translation task, 100K steps for the base model and 300K steps for big model respectively. The batch size is set as 4096.

Our model is implemented based on tensor2tensor\footnote{https://github.com/tensorflow/tensor2tensor}, which is an open source toolkit built by Tensorflow. Adam \cite{DBLP:journals/corr/KingmaB14} is chosen as default optimizer with $\beta_1 = 0.9$, $\beta_2=0.98$ and $\epsilon=10^{-9}$. Following Vaswani \shortcite{46201}, we adopt a varied learning rate over the training procedure:
\begin{equation}
\mathrm{lr = d_{model}^{-0.5} \cdot \min{(steps^{-0.5}, steps \cdot warm\_steps^{-1.5})}}
\end{equation}
where $\mathrm{steps}$ is the current training steps, and $\mathrm{warm\_steps}$ is the warmup training steps.

During the decoding step, we set beam size as 4 and length penalty $\alpha=0.6$. The maximum decoding length is set as the input length plus 50. We report the results by averaging the last 5 and 20 checkpoints for the base and big model respectively which is consistent with Vaswani \shortcite{46201}.

\begin{table*}
	\begin{subtable}[h]{0.33\textwidth}
		\centering
		\begin{tabular}{@{}l r rr@{}}
			\toprule
			Method && BLEU & Params \\
			\cmidrule{1-1} \cmidrule{3-4}
			Baseline &&31.27& 11.9M \\
			\cmidrule{1-1} \cmidrule{3-4}
			Sum & & 32.07 & -\\
			Concat & & 32.13 & +524K \\
			Gated Sum & & \textbf{32.28} & +32K \\
			\bottomrule
		\end{tabular}
		\caption{IWSLT2014 German-English (small)}
	\end{subtable}
	\begin{subtable}[h]{0.33\textwidth}
		\centering
		\begin{tabular}{@{}l r rr@{}}
			\toprule
			Method && BLEU & Params \\
			\cmidrule{1-1} \cmidrule{3-4}
			Baseline && 27.3 & 66.8M \\
			\cmidrule{1-1} \cmidrule{3-4}
			Sum && 27.4 & - \\
			Concat && 27.7 & +6.29M \\
			Gated Sum && \textbf{27.9} & +383K \\
			\bottomrule
		\end{tabular}
		\caption{WMT14 English-German (base)}
	\end{subtable}
	\begin{subtable}[h]{0.33\textwidth}
		\centering
		\begin{tabular}{@{}l r rr@{}}
			\toprule
			Method && BLEU & Params \\
			\cmidrule{1-1} \cmidrule{3-4}
			Baseline &&28.4& 213.4M \\
			\cmidrule{1-1} \cmidrule{3-4}
			Sum &&28.2 & -\\
			Concat &&28.7 & +25.1M\\
			Gated Sum && \textbf{28.8} & +1.57M \\
			\bottomrule
		\end{tabular}
		\caption{WMT14 English-German (big)}
	\end{subtable}
	\caption{\label{fuse}Experiments about fusion methods on different translation tasks. ``-" means do not cause any other parameters and ``+(number)" means the additional parameters in contrast to the baseline system.}
\end{table*}

\section{Results}
In this section, we will conduct extensive analysis to evaluate our models in terms of learning, the effect of branches and the combination, and visualization analysis. We report our results by \emph{BLEU} \cite{DBLP:conf/acl/PapineniRWZ02} scores.

\subsection{Ablation Studies}
In order to explore the performance of different SANs, we conduct a series of experiments on IWSLT14 German-English translation task. To evaluate the ability of our approach in extracting temporal order information, we also take experiments with disabling positional embedding. Table \ref{branch} summarizes our results. For the sake of fairness, all experiments adopt gated sum as the combination method for our ablation studies. As shown in Table \ref{branch}, we have the following observations:

\begin{itemize}
	
	\item When compared with $base$, experiments $a_{1-3}$ and $b_{1-4}$ can achieve at most $0.65$ BLEU gain. The improvements show that DiSAN and LSAN can obtain benefits on the encoder. When compared $d_{1-3}$ with $base$, LSAN on decoder only achieves at most $0.16$ BLEU gain. More specifically, both directional SANs are useful for the model while it seems unnecessary to combine many LSAN branches since $b_4$ does not bring the most performance.
	
	\item Comparing $c_5$ with $base$, model achieves $0.89$ BLEU gain by integrating bi-DiSAN and LSAN on the encoder. In addition, we notice that the performance almost expresses no difference between $e_1$ and $e_2$ settings with nearly $1.0$ BLEU gain.
	
	\item An interesting founding is that our method can give over 15 points improvements than $base$ model when disabling positional embedding. This phenomenon indicates that global SAN can not extract enough temporal order information. When compared experiment groups $\{a_{1-3},b_{1-4},c_{1-5}\}$ with $d_{1-4}$, we find that encoder is more sensitive than decoder when disabling positional information. We guess because decoder can be viewed as a forward SAN that means it has covered some degree of temporal order information.
	
\end{itemize}
In order to ensure consistency and save computation, our next experiments mainly adopt $e_1$ as the default setting.

\begin{table}
	\centering
	\begin{tabular}{l c}
		\toprule
		Model & BLEU \\
		\midrule
		ByteNet \cite{DBLP:journals/corr/KalchbrennerESO16} & 23.75 \\
		GNMT + RL \cite{DBLP:journals/corr/WuSCLNMKCGMKSJL16}& 24.60 \\
		ConvS2S \cite{gehring2017convs2s} & 25.16 \\
		MOE \cite{DBLP:journals/corr/ShazeerMMDLHD17} & 26.03 \\
		\midrule
		GNMT + RL Ensemble & 26.30 \\
		ConvS2S Ensemble  & 26.36 \\
		\midrule
		Transformer (base) & 27.3 \\
		HySAN (base) & \textbf{27.9} \\
		\midrule
		Transformer (big) & 28.4 \\
		HySAN (big) & \textbf{28.8} \\
		\bottomrule
	\end{tabular}
	\caption{\label{en-de}Results on WMT14 English-German translation task.}
\end{table}

\subsection{Effect of Fusion Method}
To understand which fusion method will affect model performance to what extent, we conduct some experiments to survey the influence of three alternatives. The results are reported in Table \ref{fuse}. In order to further manifest the robustness of different branches, we perform experiments on different scale model settings. From Table \ref{fuse}, we have the following summaries:
\begin{itemize}
	\item Sum fusion does not need any parameters. However, owing to the lack of parameters, sum fusion can be regarded as a linear combination. Hence, the model with sum fusion will manifest some inadaptabilities when facing with distinct model settings and different translation tasks. We can see that it only achieves $0.80$ points improvement in IWSLT translation task, and even cannot yield benefits in WMT translation task.
	
	\item Concatenation is equivalent to dense connectivity. Therefore, it will be required to perform a linear projection to match the dimensions. Moreover, concatenation shows better adaptability and flexibility than sum in performance. However, concatenation demands approximately over 10\% additional parameters and still has slight inferiority when compared with gated sum method.
	
	\item Gated sum is our preferred method. This method is more adaptive as it achieves the better performance in three settings than all of the other fusion methods. In addition, our approach also has advantage in the number of parameters since it only requires not more than 1\% additional parameters. We use this fusion to evaluate our method on the large dataset.
\end{itemize}

\begin{table}
	\centering
	\begin{tabular}{l c}
		\toprule
		Model & BLEU \\
		\midrule
		SogouNMT \cite{Wang2017SogouNM} & 26.40 \\
		XMUNMT \cite{Zhixing17nmt} & 23.40 \\
		Uedin-NMT \cite{uedin-nmt:2017} & 23.60 \\
		\midrule
		Transformer (big)  & 24.20 \\
		HySAN (big) & \textbf{25.27} \\
		\bottomrule
	\end{tabular}
	\caption{\label{zh-en} Results on WMT17 Chinese-English}
\end{table}

\subsection{Results on English-German}
Table \ref{en-de} presents the results of English-German translation. We compare our model with other various systems including RNN-based model, CNN-based model and original self-attention-based model. For the sake of fairness, we mainly compare our model with Transformer. 

Note that Transformer has achieved a strong baseline over other state-of-the-art systems. For the base model setting, our model can achieve an improvement with 0.6 BLEU points. For the big model setting, our model can also achieve 0.4 BLEU points improvement. These results suggest that our method is more competitive in different model scales. Our method is easy to implement and extend. Besides, our method only increases less than 1\% additional parameters of the original model and will not produce large redundant computation during the training phase. Above properties manifest the superiority of our method.

\subsection{Results on Chinese-English}
Table \ref{zh-en} gives the results on WMT17 Chinese-English. Note that SogouNMT is an ensemble system with rescoring and named entity techniques. From Table \ref{zh-en}, we find the performance of Transformer has surpassed all of advanced single models. However, our proposed method can still achieve nearly 1.07 BLEU improvement than the Transformer baseline. These marginal improvements also indicate the effectiveness of our approach.

\subsection{Learning Curves}
\begin{figure}
	\centering
	\includegraphics[width=0.88\columnwidth, height=38ex]{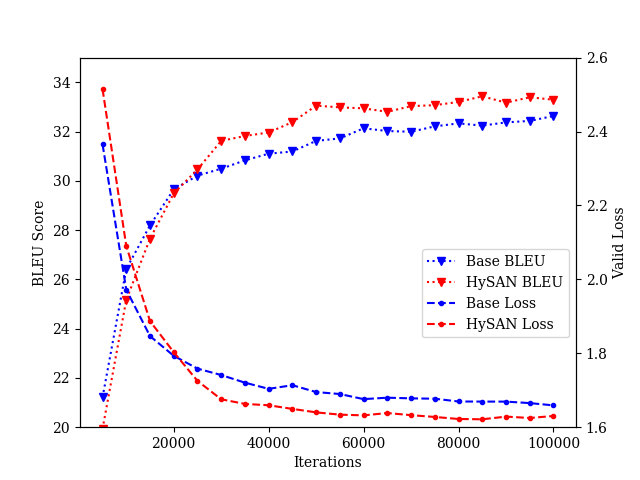}
	\vspace{-1.0em}
	\caption{\textbf{Learning curves} - translation performance (left $y$-axis) and loss function (right $y$-axis) on development set as training progresses on IWSLT14 German-English. \label{curve}}
\end{figure}

We investigate the convergence speed between our method and baseline. Our method adopts $e_1$ setting and gated sum as the fusion method. Figure \ref{curve} demonstrates the learning curves. Our model starts up slowly at the first 20K iterations during the training. However, it is pleasant to observe that our method  learns faster and surpasses the original method in BLEU score after 20K steps. Our model tends to convergence after 60K iterations and achieves a peak performance at nearly 85K steps.

\subsection{Attention Visualization}
\begin{figure}
	\centering
	\includegraphics[width=0.95\columnwidth, height=27ex]{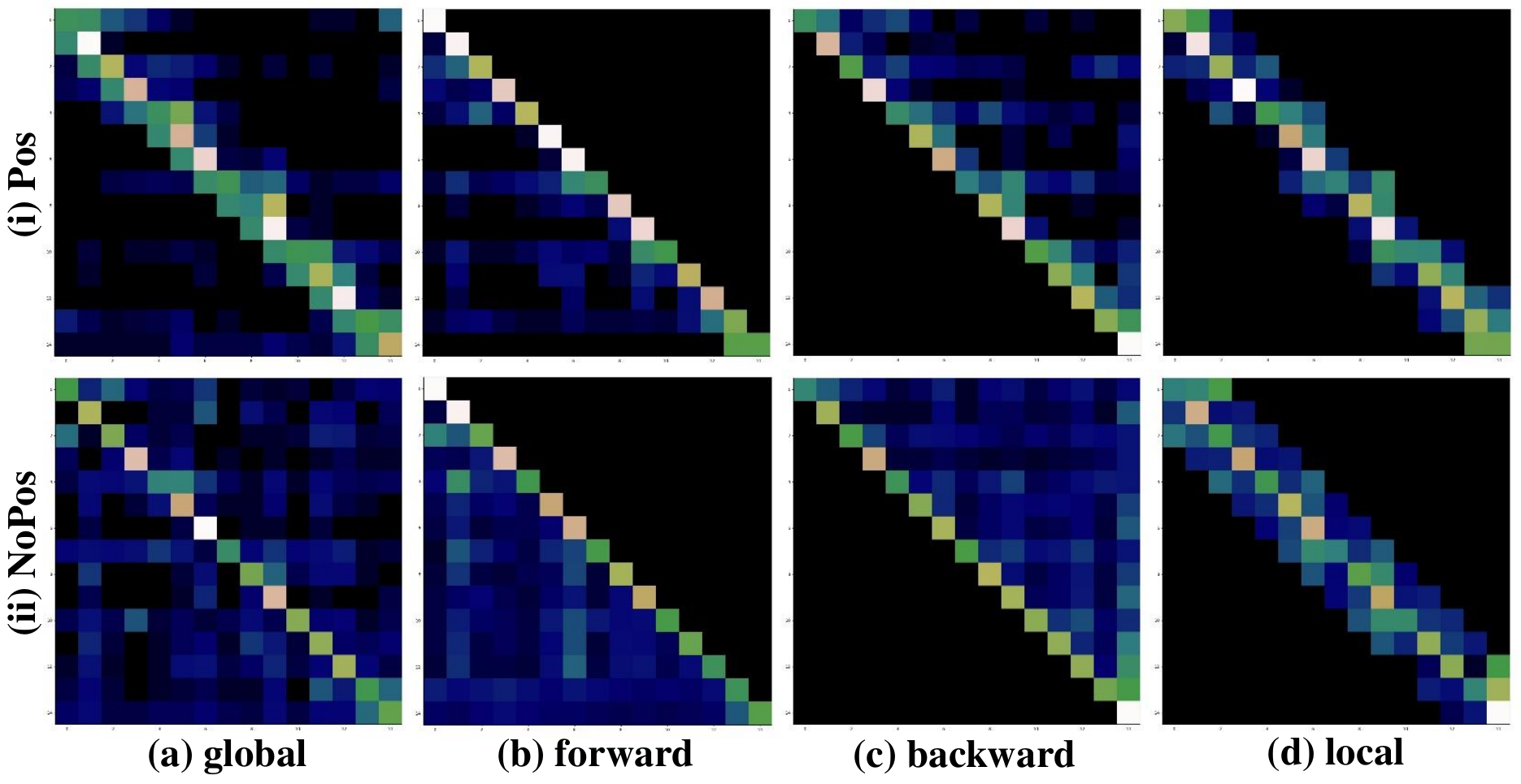}
	\caption{\textbf{Attention Visualization} of SANs in the encoder on IWSLT14 German-English. The columns represent the type of SANs. The rows represent whether using positional information.\label{atte}}
\end{figure}

To further understand the ability of our approach in extracting temporal order information, we randomly sample some sentences and visualize their attention weight of each SAN branch in the encoder of Transformer. Figure \ref{atte} presents us a detailed visualization map. We find that attention weights exhibit scattering when disabling positional information. These phenomena are existed in all of different SAN branches. We notice that attention weights of global SAN with positional information are focused on each word's surrounding. This weight distribution is also satisfied with the motivation of our proposed local SAN that partial information is essential to model dependency.  

\section{Related Work}
Inspired by Vaswani \shortcite{46201}, Shen \shortcite{DBLP:journals/corr/abs-1709-04696} introduced a variant of self-attention network for natural language inference. They decompose the global SAN into three parts, which are diag-SAN, forward-SAN and backward-SAN respectively. Different from their work, we propose another variant of SAN which can extract partial information from local context. Besides, their work uses different parameter-untied SANs for each positional mask and then fusion while in our work, all of our SANs are shared with the same alignment matrix which is more efficient. 

Ahmed \shortcite{weighted_transformer} proposed an architecture called Weighted Transformer which learned a concatenation weight to organize each attention head. Its motivation is to improve the Transformer structure for more efficient computation while our approach is to extract temporal order information and partial information. Beyond that, our method is also compatible with weighted Transformer. That means our approach can be deployed with Weighted Transformer in parallel.

\section{Conclusion}
In this paper, we propose a variant of the self-attention network named ``Hybrid Self-Attention Network (HySAN)" for neural machine translation. Our model leverages the flows from different channels to make up the disadvantages of SAN in abstracting temporal order information and partial information, and achieves comparable performance only with a few additional parameters. A squeeze gate network has been applied to our module for improving the relevance of different branches. Experimental results indicate that our model can obtain the outstanding result in multiple translation tasks over the previous method.

In the future, we will continue to study the structure and interpretations of the self-attention network in order to develop a wider range of applications for translation tasks.

\newpage
\bibliographystyle{aaai}
\bibliography{ed_res}
\end{document}